\documentclass[lettersize, journal]{IEEEtran}
\usepackage{amsmath,amsfonts}
\usepackage{algorithmic}
\usepackage{algorithm}
\usepackage{array}
\usepackage[caption=false,font=normalsize,labelfont=sf,textfont=sf]{subfig}
\usepackage{textcomp}
\usepackage{stfloats}
\usepackage{url}
\usepackage{verbatim}
\usepackage{graphicx}
\usepackage{cite}
\usepackage{booktabs}
\usepackage{hyperref}
\usepackage{svg}
\begin{document}

\title{JigsawHSI: a network for Hyperspectral Image classification}

\author{Jaime Moraga
\thanks{(\textit{Corresponding author: Jaime Moraga}, \href{mailto:jmoraga@mines.edu}{jmoraga@mines.edu})}
\thanks{J. Moraga is with the Department of Mining Engineering, Colorado School of Mines, 1610 Illinois St., Golden, CO 80401, USA }
\thanks{Document created June 2, 2022; updated Jan 18, 2025}}


\IEEEpubid{~\copyright~2022-2025 Jaime Moraga. Personal use is permitted; republication/redistribution requires the author's permission}

\maketitle

\begin{abstract}
This article describes Jigsaw, a convolutional neural network (CNN) used in geosciences and based on Inception \cite{szegedy_inception-v4_2017} but tailored for geoscientific analyses. Introduces JigsawHSI (based on Jigsaw) and uses it on the land-use land-cover (LULC) classification problem with the Indian Pines, Pavia University and Salinas hyperspectral image data sets. The network is compared against HybridSN\cite{roy_hybridsn_2020}, a spectral-spatial 3D-CNN followed by 2D-CNN that achieves state-of-the-art results on the datasets. This short article proves that JigsawHSI is able to meet or exceed HybridSN's performance in all three cases. It also introduces a generalized Jigsaw architecture in $d$-dimensional space for any number of multimodal inputs.
Additionally, the use of Jigsaw in geosciences is highlighted, while  the code and toolkit are made available. 
\end{abstract}

\begin{IEEEkeywords}
Hyperspectral Image Classification, Convolutional Neural Network, Remote Sensing, JigsawHSI, Pavia University, Indian Pines, Salinas
\end{IEEEkeywords}

\section{Introduction}
\IEEEPARstart{H}{yperspectral} image (HSI) classification is a classical task for remote sensing and machine learning practitioners, it consists in classifying the pixels from a hyperspectral image (HSI) into classes based on a given ground truth. For this tasks, several freely available data sets have been released, including Salinas Valley, Pavia University \cite{gamba_collection_2004}, and Indian Pines \cite{baumgardner_220_2015}. 

Image classification, or semantic segmentation, is a machine learning task that has been tackled extensively in the literature, its application to HSI is an interesting problem because it is a difficult task for machines that humans can do. This allows for manual labeling of the images to create a ground truth, which makes the availability of data sets for supervised learning possible.

For machines, the task is complex because of the high dimensionality of HSI, and the spatial and spectral characteristics of the classification problem. This makes na\"{\i}ve approaches to the problem to be subpar, not achieving good results.

To solve this problem, many approaches exist in the literature, with hundreds of publications in 2022 alone. The first problem is high dimensionality, so different pre-processing algorithms have been proposed, for example dimensionality reduction by using decomposition functions like Principal Component Analysis (PCA), Factor Analysis (FA), Single Value Decomposition (SVD), and others \cite{meng_lightweight_2022}. Pre-processing steps also include the application of wavelet functions, Fourier transforms \cite{miclea_local_2020} and others.

\IEEEpubidadjcol

The task of image classification and semantic segmentation has been studied in machine learning and image analysis for decades. There are two main ways to approach the problem, by using pixel-based methods or area-based methods. In pixel-based methods, each pixel is classified independently of the surrounding pixels, this has the drawback of missing all the spatial and only analyzing the spectral information.
A competing approach has been the use of both spatial and spatial information by analyzing cubes of data as a whole.

\begin{figure}[htb]
    \centering
    \includegraphics[width=2.5in]{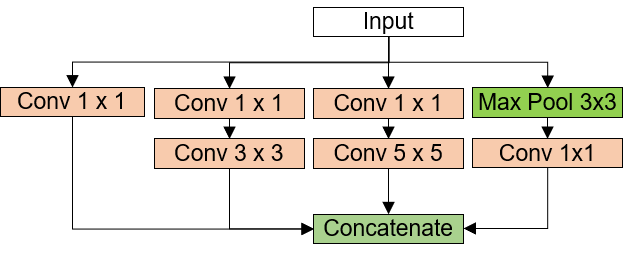}
    \caption{Inception architecture}
    \label{fig:Inception}
\end{figure}

Since the seventies, neural networks have shown that they are specially well suited for this type of problem. Since the neocognitron \cite{fukushima_neocognitron_1980}, various shallow and deep neural and convolutional neural networks (CNN) have been used, with great success, to classify images with different number of bands (or channels) of data. In general, these (for example AlexNet \cite{krizhevsky_imagenet_2017}, ResNet \cite{he_deep_2016} and GoogLeNet \cite{szegedy_going_2015}) have been limited to the red, green and blue (RGB) bands of visible light, but there is no reason to limit the analysis to just those three bands. GoogLeNet's inception module, for example, analyzes data across bands independent of depth (Figure \ref{fig:Inception}).

Among the CNN approaches applied to multispectral and hyperspectral data are 2-dimensional CNN and, lately, 3-dimensional neural networks. There have also been attempts to use hybrid approaches, as in the case of the HybridSN \cite{roy_hybridsn_2020} which uses 3-dimensional kernels to both reduce dimensionality of the input and capture spectral information, and 2-dimensional kernels to extract spatial information (e.g. image textures).

Another network proposed for general multispectral and multi-variate data is the Jigsaw network, first used to evaluate the environmental impact of an iron mine dam collapse in Brazil \cite{moraga_monitoring_2020}. This network has also been used successfully in identifying geothermal potential of two sites in Nevada (Brady and Desert Peak) \cite{moraga_geothermal_2022}. This network is capable of identifying patterns both across the channel and spatial dimensions of an image.

This short article examines the use of a variant of the Jigsaw network, the JigsawHSI network, that is highly configurable in its hyper-parameters and depth to deal with a variety of inputs. The questions to answer are whether the network can tackle HSI classification problems, whether the results are comparable to more complex hybrid or 3-dimensional approaches, what dimensionality reduction functions can be used to achieve competitive results and what hyper-parameters are relevant to achieve such results. The network, configuration routines and sample configurations are made available to the public. 

\section{The Jigsaw network}
Inception (depicted in Figure \ref{fig:Inception}) is a network that was proposed by \cite{szegedy_going_2015}, further refined in \cite{szegedy_rethinking_2016} and \cite{szegedy_inception-v4_2017}, and had 2 main objectives: a) to obtain a convolutional structure for a translation invariant vision network; and b) to introduce sparsity by using the Network-in-network idea in \cite{lin_network_2014}, that is, using multilayer perceptron (MLP) implemented as 1x1 convolutions to reduce the dimension of the inputs or outputs and thus reduce the number of operations of the network while introducing a network that captures non-linearity. 

The network has been used successfully for image classification in implementations like GoogLeNet \cite{szegedy_going_2015}. For HSI classification, several attempts have been made to use inception as a module, including recently the 3D AI \cite{fang_hyperspectral_2022} who uses two 3D inception modules, one for spatial classification and one for spectral in each layer.

\begin{figure}[!htb]
    \centering
    \includegraphics[width=3in]{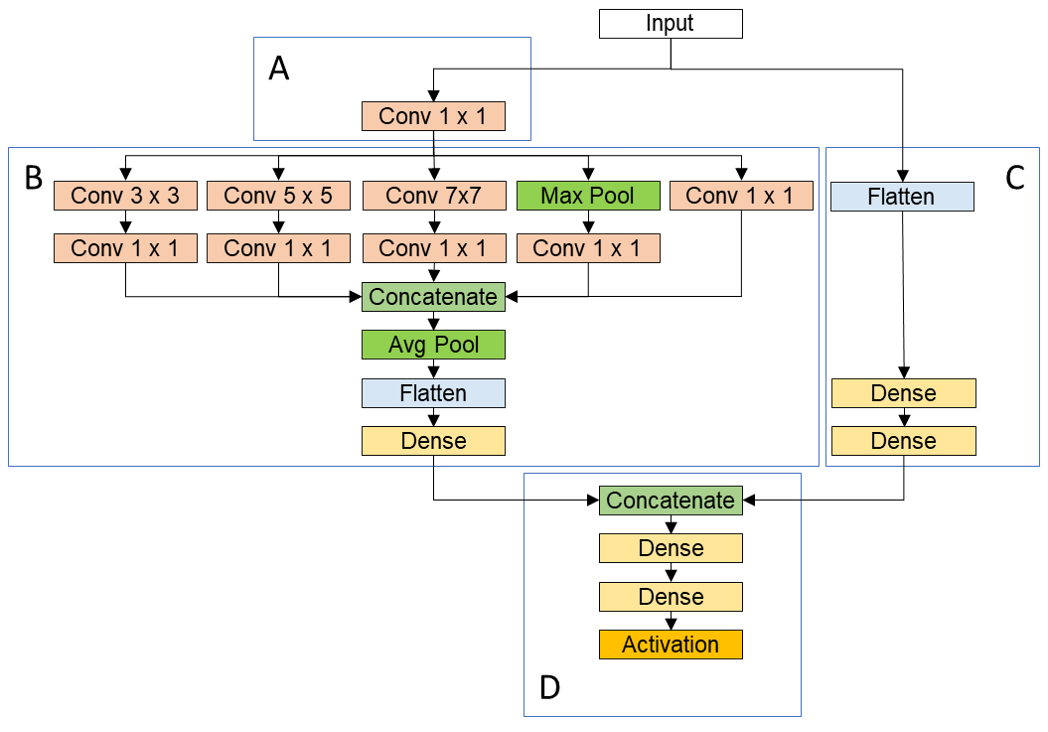}
    \caption{Jigsaw's original architecture, adapted from \cite{moraga_monitoring_2020}}
    \label{fig:Jigsaw}
\end{figure}

We proposed Jigsaw in 2019 \cite{moraga_jigsaw_2019} to address Land-use land-cover (LULC) problems with multispectral images as inputs. The core of the network is an Inception-like module. The original jigsaw network is shown in Figure \ref{fig:Jigsaw}.

The Jigsaw network’s design had the same initial 2 considerations as Inception, a) the visual representation that is translation independent (for spatial analysis), and b) introduce sparsity to allow to go deeper if required. An additional 2 considerations for this network are c) to capture the spectral information and d) not to lose resolution. For this purpose, the \textbf{A and C modules} in Figure \ref{fig:Jigsaw} were introduced. The \textbf{A module} uses the Network-in-network concept of \cite{lin_network_2014}: an MLP to capture linear relationship among bands in the HSI, this module can contain one or more  $(1 \, \times \, 1)$ convolutions in series. The \textbf{C module} acts as a ResNet \cite{he_deep_2016} -- making the initial image available for classification in deeper layers --, also highlighting both spatial and spectral information. This \textbf{C module} can increase the size and complexity of the network substantially  –- especially for hyperspectral images where the number of bands can be in the hundreds -–, so there are two ways to reduce that impact, the first is to crop the image keeping just the center pixel (which is the determinant for the network’s classification), and then sending that pixel’s spectrum through the dense connections; the second way is to preprocess the input to reduce the number of bands by doing dimensionality reduction with algorithms like principal component analysis (PCA) or factor analysis (FA).

Finally, \textbf{module D} concatenates the results of the previous layers and feeds two Dense layers finalizing in a Softmax activation function to produce the resulting classes.

\begin{figure}[!hbt]
    \centering
    \includegraphics[width=3in]{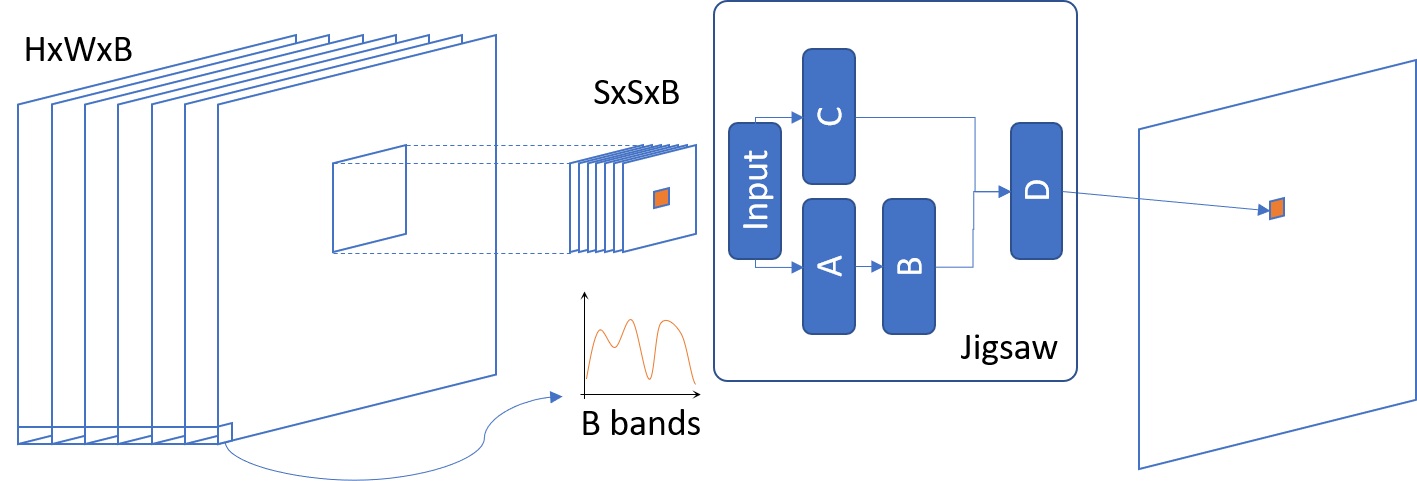}
    \caption{Jigsaw's process}
    \label{fig:Jigsaw_process}
\end{figure}

The whole process is depicted in Figure \ref{fig:Jigsaw_process}. The first step is to acquire an HSI, the initial dimensions will be called H (height), W (width), and B (bands), to define a figure of dimensions $(H \, \times \, W \, \times \, B)$, representing pixels $p_{i,j,k}$ in $\mathbb{R}^3$. From this image, each $(1 \, \times \, 1 \, \times \, B)$ slice represents the profile of the spectrum for that pixel, that is, the intensity at each bandwidth represented by each band in B. From this image, tiles of side S are extracted, to form $(H \, \times \, W)$ tiles of dimensions $(S \, \times \, S \, \times \, B)$.

These $(S \, \times \, S \, \times \, B)$ tiles (the “jigsaw pieces”) are processed by the Jigsaw network and the results of training and prediction are mapped back to an $(H \, \times \, W \, \times \, 1)$ image (“completing the jigsaw puzzle”).
To reduce overfitting, L2 kernel regularization is added at each layer, and 40\% dropout layers are added after each dense layer. 
The additional purpose of having \textbf{module A} is to create a single set of filters from the spectral information, to present them to the spatial analysis module. The logic of adding a $(7 \, \times \, 7)$ filter to the Inception-like \textbf{module B} is to capture more of the spatial information around the center pixel. This, knowing that the pixels in the surroundings are spatially correlated with the center pixel. In its first application, this corresponded to a region of radius 35 meters from the center pixel, that matched our spatial statistics analysis of the image.

In summary, the originality of Jigsaw lies in: spatial-spectral focus for multispectral and hyperspectral data by adding specific modules that handle the channels or bands with MLPs; use of inception-like module with bigger filters to manage spatially correlated pixels around the center pixel; use of spatial statistics to inform the number and size of the additional filters in module \textbf{B}; and the use of Tikhonov (L2) kernel regularization in convolutional layers, and dropout after dense layers, to limit the overfitting of the network. This last point may reduce overall accuracy in the training set, but increases generalization and thus accuracy on the test sets.

\section{The JigsawHSI network}
\begin{figure*}[!htb]
    \centering
    \includegraphics[width=6in]{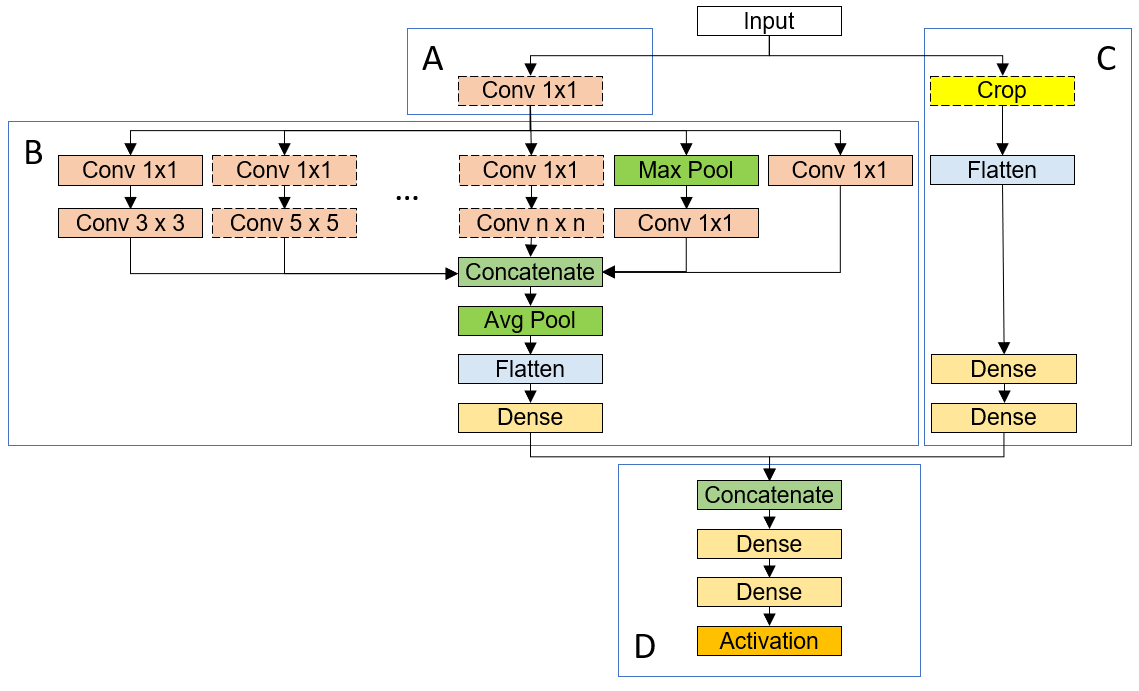}
    \caption{JigsawHSI architecture. The dotted lined layers are optional}
    \label{fig:JigsawHSI}
\end{figure*}

The JigsawHSI network (Figure \ref{fig:JigsawHSI}), is an adaptation of two networks based on the original Jigsaw network. 
The first network was used to determine the surface impact of a tailings dam collapse in Brumadinho, Brazil by classifying multispectral pixels from Sentinel-2 in before and after pictures, identifying the areas affected by the dam's iron ore waste tailings \cite{moraga_monitoring_2020}. The second application was in the Geothermal AI where, by using multi-variate input layers (Mineral Markers, Temperature and Faults) that were the result of machine learning preprocessing, the network classifies pixels in an image as having or not geothermal potential \cite{moraga_geothermal_2022}.  In both cases, the inception network \cite{szegedy_inception-v4_2017} was adapted to better capture the correlation in surface anomalies by increasing the number of internal filters of the inception network and adding a first 1x1 convolution after the input layer to better manage relationships across bands.

The JigsawHSI's architecture builds upon this idea by both varying the number of internal $(n \, \times \, n)$ filters and making the first $(1 \, \times \, 1)$ convolutional layer optional, as shown in Figure \ref{fig:JigsawHSI}.

The network uses two parallel blocks (\textbf{modules A and B} on the left and \textbf{module C} on the right) that are merged in the second concatenate layer. The left side block uses the Inception-like module to capture spatial and spectral information, increasing in \textbf{module B} the number of $(n \times n)$ filters from the same block in Jigsaw (Figure \ref{fig:Jigsaw}), finishing with a concatenate and average pooling layers; while the right in \textbf{module C} introduces a Crop layer that only uses the central pixel of the image, this module then uses a flattening layer to discard spatial information and analyze all the spectral information of each pixel looking for linear and non-linear relationships. The bottom of the  network (\textbf{module D}) has two fully connected dense layers and uses Softmax as the activation function, turning the results into classes as in the case of the vanilla version of Jigsaw.

 This network's code and Jupyter Notebook are available in \url{https://github.com/jmoraga-mines/JigsawHSI}. The network is highly flexible, and can be configured by editing the {\texttt{config.ini}} file in the root directory.  

The configuration includes the ability to define an analysis window that will partition the HSI $(width, height, bands)$ in squares of side $S$, creating cubes for analysis by the network of dimensions $(S, S, bands)$, a decomposition function can also be specified to reduce the dimensionality to $c$ channels. The final input to the network will be of dimensions $(S, S, c)$. Optionally, a layer with $(1 \times 1)$ convolutions can be added to perform filtering in the dimension of the channels (as in \cite{lin_network_2014}), this starts the analysis in the spectral dimension of the HSI by the neural network, in these cases, the new value of $c$ will be the number of filters defined for this step. 

For training, several hyperparameters can be defined, these include the optimizer, learning rate of the optimizer, batch size, maximum number of epochs and a patience parameter to determine the number of epochs with no improvement required for early stopping of the training.

\section{Configuration and training}
The configuration parameters for training are in Table \ref{tab:parameters} for each of the datasets: Indian Pines (IP), Pavia University (PU), and the Salinas scene (SA) .

\begin{table}[htb]
    \centering
    \caption{Experimental parameters for each JigsawHSI result}
    \label{tab:parameters}
    \begin{tabular}{r|c|c|c}
    \toprule
         & \multicolumn{3}{c}{Values}   \\
        Parameters \hfill & IP & PU & SA   \\
        \midrule
        Window size & 27 & 25 & 25   \\
        Decomposition & FA & SVD & FA   \\
        Input channels & 9 & 9 & 12   \\
       \multicolumn{4}{l}{Network design:}  \hfill \\
        HSI Filters & None & 512 & None   \\
        Filter size & 9 & 9 & 7   \\
        \multicolumn{4}{l}{Hyperparameters:} \hfill  \\
        Optimizer & Adadelta & Adadelta & SGD   \\
        Learning rate & 0.1 & 0.1 & 0.01   \\
        Batch size & 106 & 120 & 132   \\
        Max. Epochs & 500 & 500 & 500   \\
        Max. Patience & 20 & 40 & 20   \\
    \bottomrule
    \end{tabular}
\end{table}

Depending on the dataset, windows of different sizes can be optimal, in this case, IP uses a $(27\times 27)$ window, while PU and SA use $(25\times 25)$.  

To simplify the analysis of the network, 4 decomposition functions can be used for dimensionality reduction: Principal Component Analysis (PCA), Functional Analysis (FA), Truncated Single Value Decomposition (SVD) and Non-negative Matrix Factorization (NMF), the number of final components can also be defined. For IP, FA and 9 factors are selected; PU uses SVD and 9 values; while SA uses FA and 12 factors.

An HSI layer with 512 filters is specified for PU. Therefore, the input's shapes will be $(27, 27, 9)$ for IP, $(25, 25, 512)$ for PU, and $(25, 25, 9)$ for SA. 

The internal filters in the Inception-like network will go up to $(9 \times 9)$ for IP and PU, and $(7 \times 7)$ for SA. 

The optimizers are also flexible, accepting Stochastic Gradient Descent (SGD), Adam and Adadelta; the learning rate of these optimizers is also configurable. The configuration used is $Adadelta(0.1)$ for IP and PU, and $SGD(0.01)$ for SA.

Batch sizes are 106 for IP, 120 for PU and 132 for SA. Maximum number of epochs is 500  for all cases. Patience is 20 for IP and SA, while PU requires to wait longer, with a patience of 40.

The training was performed in a desktop computer running Ubuntu 20.4, with a single NVDIA card, Pyhton 3.8 and the latest version of CUDA.

For comparison, we used HibridSN (\url{https://github.com/gokriznastic/HybridSN}, published in IEEE GEOSCIENCE AND REMOTE SENSING LETTERS, VOL. 17, NO. 2, FEBRUARY 2020). Both networks were trained using 30\% of the samples for training and 70\% of the samples for testing. 

\section{Evaluation and Discussion}

The convolutional neural network (CNN) is one of the most used and successful networks in computer vision and visual data processing. Traditionally, these networks have been designed with 2-dimensional (2D) CNN filters due to the datasets having relatively shallow three band (RGB) images, and in this avoiding the curse of dimensionality. In the case of HSI, even after dimensionality reduction, the number of bands or channels are significantly larger than three. 

HybridSN is a hybrid spectral-spatial 3-dimensional (3D) CNN followed by a a spatial 2D-CNN \cite{roy_hybridsn_2020}. In theory, this should ease the representation of joint spatial-spectral information by the network. The HybridSN network was tested on all three datasets and results compared to five additional networks, achieving state-of-the-art results.

In Table \ref{tab:experimental_results}, we can see that the JigsawHSI obtains results that are equivalent or better than HybridSN. In the case of Indian Pines, JigsawHSI achieves lower scores in overall accuracy (OA), Cohen Kappa (Kappa) and average accuracy (AA), but in the first two cases the difference is near 0.01\% and well within the margin of error of HybridSN. 
For the Pavia University dataset, the JigsawHSI obtains better results than HybridSN, achieving a 100\% accuracy for the test set. In the case of Salinas, both networks obtain the maximum accuracy.

\begin{table*}[!htb]
    \centering
    \caption{Experimental classification metrics (in percentages) for the three datasets. OA: Overall accuracy, Kappa: Kappa factor, AA: Average accuracy}
    \label{tab:experimental_results}
    \begin{tabular}{c|c c c | c c c | c c c}
    \toprule
        Model & \multicolumn{3}{c|}{Indian Pines} & \multicolumn{3}{c|}{Pavia University} & \multicolumn{3}{c}{Salinas} \\
         & OA & Kappa & AA & OA & Kappa & AA & OA & Kappa & AA   \\
        \midrule
        HybridSN & \bf{99.75 \textpm  0.1} & \bf{99.71} \textpm  0.1  & \bf{99.63 \textpm  0.2}  & 99.98  & 99.98  & 99.97 & \bf{100} & \bf{100} & \bf{100}  \\
        \bf{JigsawHSI} & 99.74 & 99.70 & 98.11 & \bf{100} & \bf{100} & \bf{100} & \bf{100} & \bf{100} & \bf{100} \\
    \bottomrule
    \end{tabular}
\end{table*}

\begin{figure*}[!thb]
\centering
\subfloat[]{\includegraphics[width=2.3in]{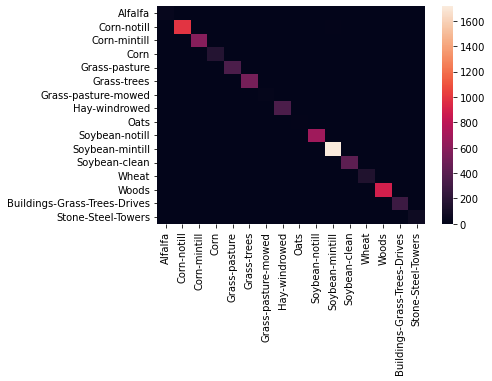}%
\label{fig:IP_heatmap}}
\hfil
\subfloat[]{\includegraphics[width=2.3in]{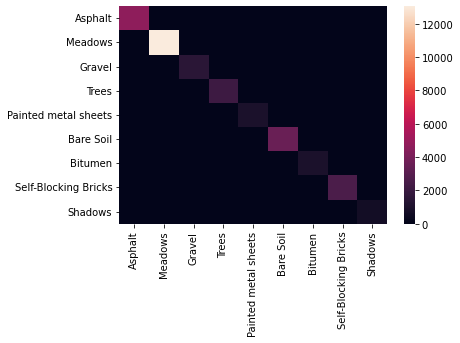}%
\label{fig:PU_heatmap}}
\hfil
\subfloat[]{\includegraphics[width=2.3in]{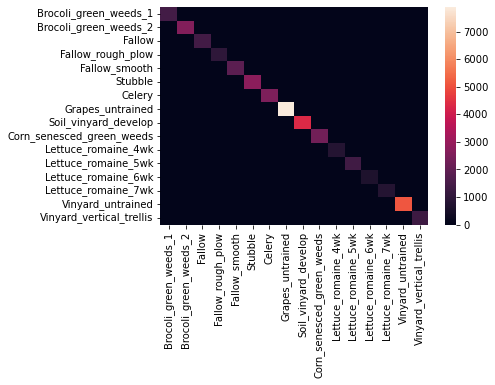}%
\label{fig:SA_heatmap}}
\caption{Confusion matrix heatmaps for (a) Indian Pines, (b) Pavia U., (c) Salinas}
\label{fig:heatmaps}
\end{figure*}

In Figure \ref{fig:heatmaps} the confusion matrix heatmaps show that in all cases the accuracy was perfect or almost perfect.

This is a surprising result at first sight because the use of 3D-CNN for spectral analysis cannot be replicated in full by the JigsawHSI. This implies that either the 3D-CNN is not needed for this case, or the 3D-CNN is not helping the spectral representation of the image. We theorize that by using dimensionality reduction as a first step in both networks, the 3D-CNN is not needed to represent in full the relationships between channels.

The other anomaly is in the AA score. This comes as a result of JigsawHSI not being able to discriminate well the oats samples in IP. This is most probably caused by the sparsity of the class, where only 14 test samples are provided.

In general, the JigsawHSI is able to match or improve on the results of the HybridSN.

\section{Applications in geoscience, a generalized Jigsaw}
The Jigsaw network was first published in \cite{moraga_monitoring_2020}, as explained in the previous section, where it was used to estimate the environmental effects of a tailings dam collapse in Brazil. In this application, two multispectral images were captured before and after the dam's collapse. When training on the "before" image and applying it to the "after" image, the network was able to generalize well achieving an 86\% classification accuracy. When training with a small subset of points of the "after" image, the network was able to generalize to the rest of the image with a 98\% accuracy. The network performed especially well when discriminating the Mine and Tailings class, with a 98.4\% accuracy, being this the most critical class to determine the impact to populated and unpopulated areas \cite{moraga_monitoring_2020}.

\begin{figure}[!htb]
    \centering
    \includegraphics[width=3in]{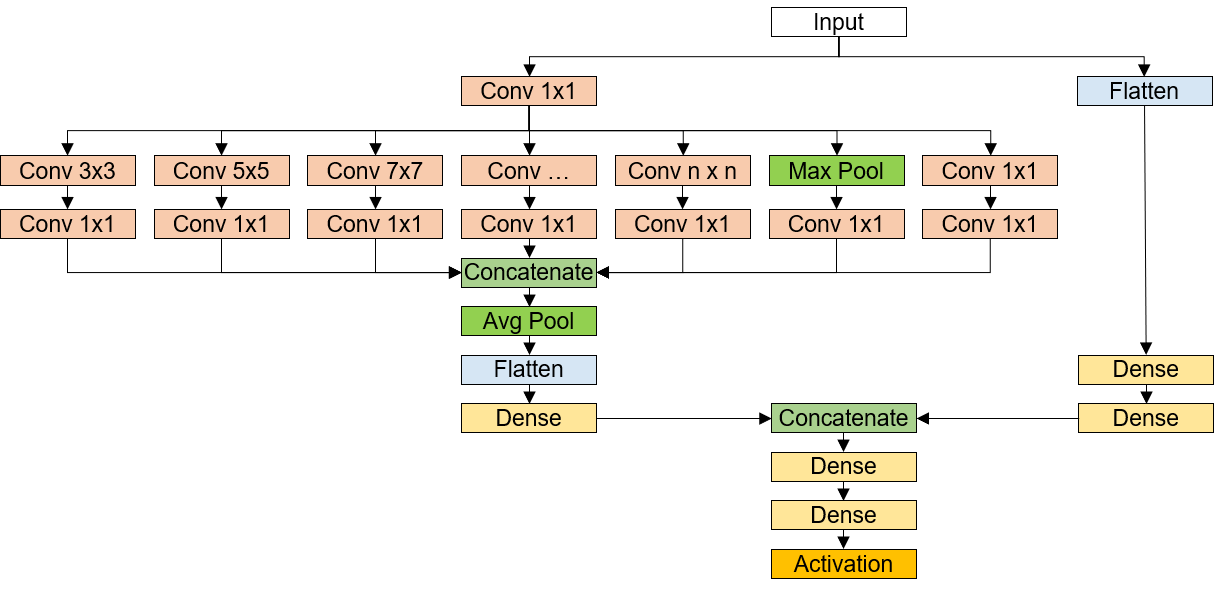}
    \caption{The Geothermal AI}
    \label{fig:GeoAI}
\end{figure}

The second application in geosciences was the use of the same concepts in geothermal exploration. In \cite{moraga_geothermal_2022}, we introduced the Geothermal AI (Figure \ref{fig:GeoAI}).

By using the Geothermal AI for exploration we were able to obtain robust results in classification on two sites, the Brady and Desert Peak geothermal sites in Churchill County, NV. The network has unique characteristics in that the input was not a multispectral or hyperspectral image but a multi-modal series of bands that were the results of preprocessing with other ML algorithms or geological/geophysical information. In this paper, we also introduced a semi-automated labeling process to create the outputs required for supervised training of the Geothermal AI, and described the preprocessing used to create input layers to the AI.

The Geothermal AI was able to achieve 92-95\% accuracy in the training sets, and 72-76\% accuracy when applied to the opposite site, showing promising ability to generalize. The network was able to highlight the areas where both operating geothermal plants were located in each case.

In this paper, we show a third application in which the network shows promising results to classify HSI images for LULC. And we believe this network can be used in other geoscience classification applications given its ability to discriminate channel information and generalize spatially.

\begin{figure}[!htb]
    \centering
    \includegraphics[width=3in]{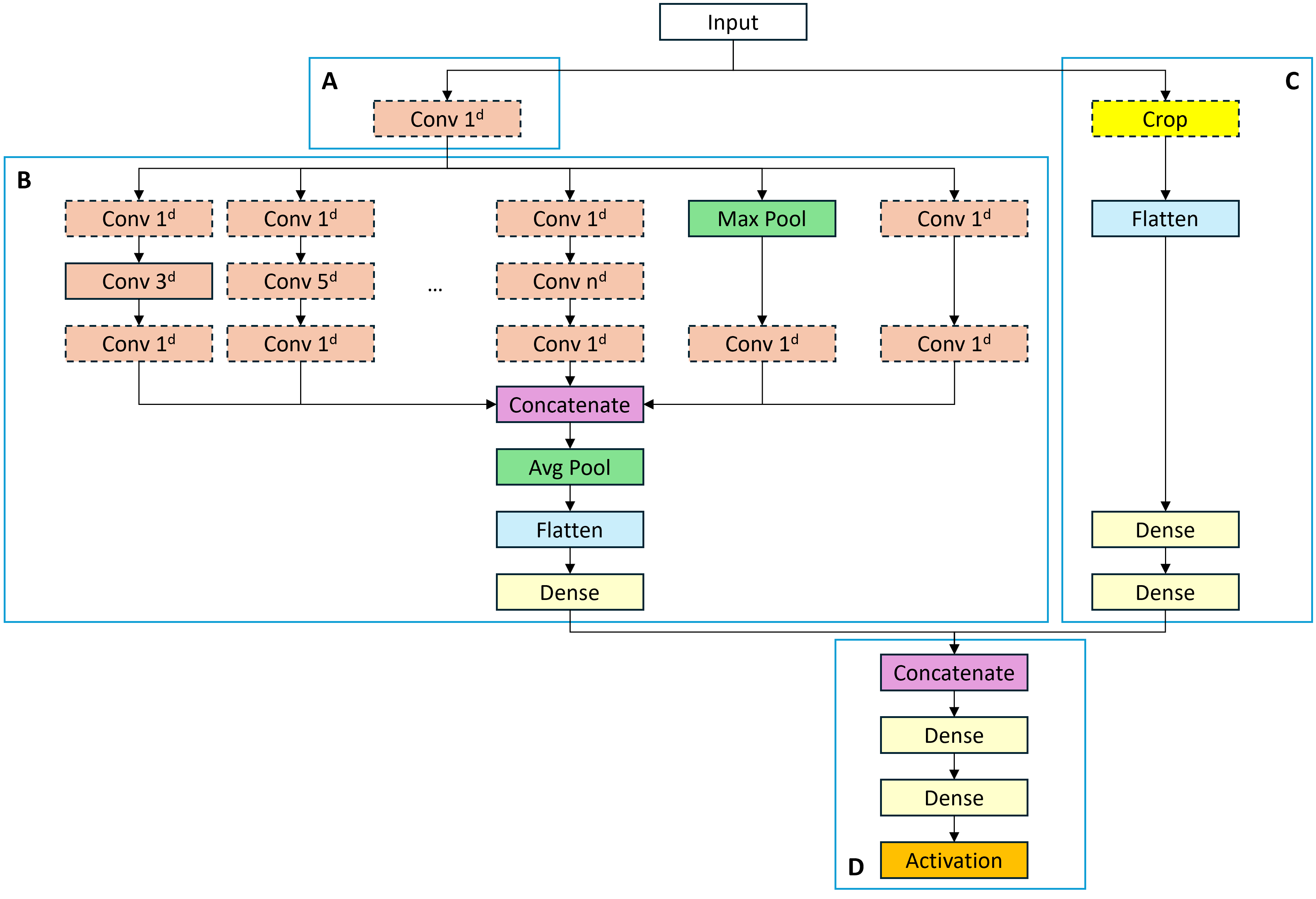}
    \caption{The generalized Jigsaw network}
    \label{fig:Jigsaw_general}
\end{figure}

A generalized network is shown in Figure \ref{fig:Jigsaw_general}. This generalized network is the result of merging JigsawHSI and Geothermal AI. The Input layer can be multispectral, hyperspectral, multi-modal or the result of the application of dimensionality reduction algorithms to any $d$-dimensional application where topological correlation exists (e.g. time series of oceanic data, underground biome analysis, geographical area ecological investigation). \textbf{Module A} is an optional module that can contain one or more $(1 ^ d)$ for 2D, 3D, or $d$-dimensional convolutions. In \textbf{module B}, network-in-network can be optionally applied before and/or after the larger $(k ^ d)$ convolutions, and these convolutions can be of size $(1 ^ d)$, $(3 ^ d)$, ..., up to $(n ^ d)$ based on the $d$-dimensional spatial analysis of the Input image, this spatial analysis also informs the size of the window (S) used to create the input tiles.
In \textbf{module C}, an optional Crop layer captures only the center pixel to reduce the size of that module, while capturing non-linearity in the bands of the input image.

\section{Conclusion}

In this article, the JigsawHSI network was introduced, a slight modification of the Jigsaw network and Geothermal AI networks. The network was tested against the Indian Pines, Pavia University and Salinas Valley datasets, three freely datasets widely used in hyperspectral image classification.

The JigsawHSI network achieves results that are either equivalent or better than HybridSN under the same conditions, showing that with 2D-CNNs results match the state-of-the-art results a 3D-2D-CNN when dimensionality reduction is applied as a first step.

It has also been shown how the Jigsaw architecture performs well in geosciences when processing images for pixel by pixel classification (in land-use land-cover and geothermal exploration), and we proposed a generalized Jigsaw architecture that merges the design of JigsawHSI and the Geothermal AI.

Additional research directions include to test of the networks using smaller test samples, eliminating the dimensionality reduction and using spatial cross validation to better assess the generalization capabilities and required hyperparameters of the networks.

\section*{Acknowledgments}
I want to thank the support of my PhD thesis advisor, Prof. Dr. H. Sebnem Duzgun and the funding of the US Department of Energy, which allowed me to spend time on this foundational model for multidimensional analysis.
Hyperspectral datasets collected by M Graña, MA Veganzons, B Ayerdi and provided by Grupo de Inteligencia Computacional (GIC) (\href{https://www.ehu.eus/ccwintco/index.php/Hyperspectral_Remote_Sensing_Scenes}{https://ehu.eus/ccwintco}). Pavia scenes originally courtesy of Prof. Paolo Gamba and Prof. Fabio Dell'Acqua from the Telecommunications and Remote Sensing Laboratory, Pavia university (Italy). 
Special thanks to Gopal Krishna, for making HybridSN available in Github.

\bibliographystyle{plain}
\bibliography{references}

\begin{thebibliography}{10}

\bibitem{baumgardner_220_2015}
Marion~F. Baumgardner, Larry~L. Biehl, and David~A. Landgrebe.
\newblock 220 {Band} {AVIRIS} {Hyperspectral} {Image} {Data} {Set}: {June} 12, 1992 {Indian} {Pine} {Test} {Site} 3, September 2015.
\newblock doi:10.4231/R7RX991C.

\bibitem{fang_hyperspectral_2022}
Bei Fang, Yu~Liu, Haokui Zhang, and Juhou He.
\newblock Hyperspectral {Image} {Classification} {Based} on {3D} {Asymmetric} {Inception} {Network} with {Data} {Fusion} {Transfer} {Learning}.
\newblock {\em Remote Sensing}, 14(7):1711, January 2022.
\newblock Number: 7 Publisher: Multidisciplinary Digital Publishing Institute.

\bibitem{fukushima_neocognitron_1980}
Kunihiko Fukushima.
\newblock Neocognitron: {A} self-organizing neural network model for a mechanism of pattern recognition unaffected by shift in position.
\newblock {\em Biological Cybernetics}, 36(4):193--202, April 1980.

\bibitem{gamba_collection_2004}
P.~Gamba.
\newblock A collection of data for urban area characterization.
\newblock In {\em {IGARSS} 2004. 2004 {IEEE} {International} {Geoscience} and {Remote} {Sensing} {Symposium}}, volume~1, page~72, 2004.
\newblock Pavia Universitt dataset.

\bibitem{he_deep_2016}
Kaiming He, Xiangyu Zhang, Shaoqing Ren, and Jian Sun.
\newblock Deep {Residual} {Learning} for {Image} {Recognition}.
\newblock In {\em 2016 {IEEE} {Conference} on {Computer} {Vision} and {Pattern} {Recognition} ({CVPR})}, pages 770--778, June 2016.
\newblock ISSN: 1063-6919.

\bibitem{krizhevsky_imagenet_2017}
Alex Krizhevsky, Ilya Sutskever, and Geoffrey~E. Hinton.
\newblock {ImageNet} classification with deep convolutional neural networks.
\newblock {\em Communications of the ACM}, 60(6):84--90, May 2017.

\bibitem{lin_network_2014}
Min Lin, Qiang Chen, and Shuicheng Yan.
\newblock Network {In} {Network}.
\newblock Technical Report arXiv:1312.4400, arXiv, March 2014.
\newblock arXiv:1312.4400 [cs] type: article.

\bibitem{meng_lightweight_2022}
Zhe Meng, Licheng Jiao, Miaomiao Liang, and Feng Zhao.
\newblock A {Lightweight} {Spectral}-{Spatial} {Convolution} {Module} for {Hyperspectral} {Image} {Classification}.
\newblock {\em IEEE Geoscience and Remote Sensing Letters}, 19:1--5, 2022.

\bibitem{miclea_local_2020}
Andreia~Valentina Miclea, Romulus Terebes, and Serban Meza.
\newblock Local binary patterns and {Fourier} transform based hyperspectral image classification.
\newblock In {\em 2020 {International} {Symposium} on {Electronics} and {Telecommunications} ({ISETC})}, pages 1--4, 2020.

\bibitem{moraga_geothermal_2022}
J.~Moraga, H.~S. Duzgun, M.~Cavur, and H.~Soydan.
\newblock The {Geothermal} {Artificial} {Intelligence} for geothermal exploration.
\newblock {\em Renewable Energy}, 192:134--149, June 2022.

\bibitem{moraga_jigsaw_2019}
Jaime Moraga, Gurbet Gurkan, and Prof. Dr. H.~Sebnem Duzgun.
\newblock Jigsaw: {A} {Land} use {Land} cover classifier for multispectral images.
\newblock Technical report, Colorado School of Mines, Golden, CO, 2019.
\newblock Unpublished, not peer reviewed.

\bibitem{moraga_monitoring_2020}
Jaime Moraga, Gurbet Gurkan, and Prof. Dr. H.~Sebnem Duzgun.
\newblock Monitoring {The} {Impacts} of a {Tailings} {Dam} {Failure} {Using} {Satellite} {Images}.
\newblock In {\em {USSD} {Elevate} {Conference} 2020}, volume 2020, Denver, CO, April 2020. United States Society on Dams (USSD).

\bibitem{roy_hybridsn_2020}
Swalpa~Kumar Roy, Gopal Krishna, Shiv~Ram Dubey, and Bidyut~B. Chaudhuri.
\newblock {HybridSN}: {Exploring} 3-{D}–2-{D} {CNN} {Feature} {Hierarchy} for {Hyperspectral} {Image} {Classification}.
\newblock {\em IEEE Geoscience and Remote Sensing Letters}, 17(2):277--281, February 2020.
\newblock Conference Name: IEEE Geoscience and Remote Sensing Letters.

\bibitem{szegedy_inception-v4_2017}
Christian Szegedy, Sergey Ioffe, Vincent Vanhoucke, and Alexander Alemi.
\newblock Inception-v4, {Inception}-{ResNet} and the {Impact} of {Residual} {Connections} on {Learning}.
\newblock {\em Proceedings of the AAAI Conference on Artificial Intelligence}, 31(1), February 2017.

\bibitem{szegedy_going_2015}
Christian Szegedy, Wei Liu, Yangqing Jia, Pierre Sermanet, Scott Reed, Dragomir Anguelov, Dumitru Erhan, Vincent Vanhoucke, and Andrew Rabinovich.
\newblock Going deeper with convolutions.
\newblock In {\em 2015 {IEEE} {Conference} on {Computer} {Vision} and {Pattern} {Recognition} ({CVPR})}, pages 1--9, June 2015.
\newblock ISSN: 1063-6919.

\bibitem{szegedy_rethinking_2016}
Christian Szegedy, Vincent Vanhoucke, Sergey Ioffe, Jon Shlens, and Zbigniew Wojna.
\newblock Rethinking the {Inception} {Architecture} for {Computer} {Vision}.
\newblock In {\em 2016 {IEEE} {Conference} on {Computer} {Vision} and {Pattern} {Recognition} ({CVPR})}, volume 2016 of {\em 2016 {IEEE} {Conference} on {Computer} {Vision} and {Pattern} {Recognition} ({CVPR})}, pages 2818--2826, Las Vegas, Nevada, June 2016. IEEE.
\newblock ISSN: 1063-6919.

\end{thebibliography}

\newpage

\section{Biography Section}
\vspace{11pt}

\begin{IEEEbiography}[{\includegraphics[width=1in,height=1.25in,clip,keepaspectratio]{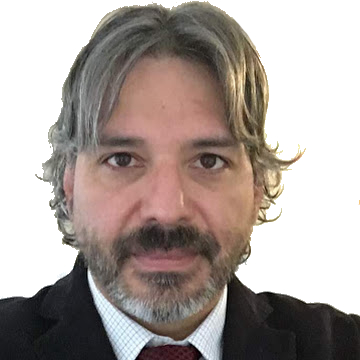}}]{Jaime Moraga} is a PhD student at Colorado School of Mines,
Golden, CO
\end{IEEEbiography}

\vspace{11pt}

\vfill
\newpage
\section{Appendixes}
{\appendix[Indian Pines results]
\begin{verbatim}
Accuracy by target (in percentages):
100.0000 : Alfalfa
 99.0000 : Corn-notill
100.0000 : Corn-mintill
100.0000 : Corn
100.0000 : Grass-pasture
100.0000 : Grass-trees
100.0000 : Grass-pasture-mowed
100.0000 : Hay-windrowed
 71.4286 : Oats
100.0000 : Soybean-notill
 99.8255 : Soybean-mintill
 99.5181 : Soybean-clean
100.0000 : Wheat
100.0000 : Woods
100.0000 : Buildings-Grass-Trees-Drives
100.0000 : Stone-Steel-Towers
\end{verbatim}
}
{\appendix[Pavia University results]
\begin{verbatim}
Accuracy by target (in percentages):
100.0000 : Asphalt
100.0000 : Meadows
100.0000 : Gravel
 99.9534 : Trees
100.0000 : Painted metal sheets
100.0000 : Bare Soil
100.0000 : Bitumen
100.0000 : Self-Blocking Bricks
100.0000 : Shadows
\end{verbatim}

}
{\appendix[Salinas Valley, CA results]
\begin{verbatim}
Accuracy by target (in percentages):
100.0000 : Broccoli_green_weeds_1
100.0000 : Broccoli_green_weeds_2
100.0000 : Fallow
100.0000 : Fallow_rough_plow
 99.9467 : Fallow_smooth
100.0000 : Stubble
100.0000 : Celery
100.0000 : Grapes_untrained
100.0000 : Soil_vineyard_develop
100.0000 : Corn_senesced_green_weeds
100.0000 : Lettuce_romaine_4wk
100.0000 : Lettuce_romaine_5wk
100.0000 : Lettuce_romaine_6wk
100.0000 : Lettuce_romaine_7wk
100.0000 : Vineyard_untrained
100.0000 : Vineyard_vertical_trellis
\end{verbatim}
}


\end{document}